\documentclass{article}

\PassOptionsToPackage{numbers, compress,square}{natbib}

\usepackage[final]{neurips_2021}

\usepackage[utf8]{inputenc} 
\usepackage[T1]{fontenc}    
\usepackage{hyperref}       
\usepackage{url}            
\usepackage{booktabs}       
\usepackage{amsfonts}       
\usepackage{nicefrac}       
\usepackage{microtype}      
\usepackage{xcolor}         

\usepackage{graphicx}
\usepackage{amsmath}
\usepackage{amsthm}
\usepackage{amssymb}
\usepackage{subcaption}
\usepackage{multirow}

\usepackage{xr-hyper}
\usepackage{enumitem}

\usepackage{hyperref}
\usepackage[capitalise]{cleveref} 
\usepackage{wrapfig,lipsum,booktabs}
\usepackage{algorithm}
\usepackage[noend]{algpseudocode}
\newtheorem*{theorem}{Lemma}

\newcommand\BCH[1]{\textcolor{cyan}{(\textbf{Brian}: #1)}}

\newcommand\revision[1]{\textcolor{black}{#1}}

\title{Accelerated Sparse Neural Training: A Provable and Efficient Method to Find N:M Transposable Masks}

\newcommand*\samethanks[1][\value{footnote}]{\footnotemark[#1]}
\author{
Itay Hubara\,${^\dagger}{^\circ}$\thanks{Equal contribution.}\quad 
Brian Chmiel\,${^\dagger}{^\circ}$\samethanks[1]\quad 
Moshe Island\,$^\dagger$\quad \\
\textbf{Ron Banner\,$^\dagger$ \quad 
Joseph (Seffi) Naor\,$^\diamond$\quad 
Daniel Soudry\,$^\circ$} 
\\[0.2cm]
$^\dagger$Habana Labs  --  An Intel company, Caesarea, Israel\\ 
$^\circ$Electrical and Computer Engineering Department - Technion, Haifa, Israel\\
$^\diamond$ Computer Science Department - Technion, Haifa, Israel
\\[0.2cm]
\small{\texttt{\{\href{mailto:ihubara
@habana.ai}{ihubara}, \href{mailto:bchmiel@habana.ai}{bchmiel}, \href{mailto:misland@habana.ai}{misland}, \href{mailto:rbanner@habana.ai}{rbanner}\}@habana.ai}}\\
\small{\texttt{\{\href{mailto:naor@cs.technion.ac.il}{naor}\}@cs.technion.ac.il}}\\
\small{\texttt{\{\href{mailto:daniel.soudry@gmail.com}{daniel.soudry}\}@gmail.com}}
}

\begin{document}

\maketitle


\begin{abstract}
Unstructured pruning reduces the memory footprint in deep neural networks (DNNs). Recently, researchers proposed different types of structural pruning intending to reduce also the computation complexity. In this work, we first suggest a new measure called mask-diversity which correlates with the expected accuracy of the different types of structural pruning. We focus on the recently suggested $N:M$ fine-grained block sparsity mask, in which for each block of $M$ weights, we have at least $N$ zeros. While $N:M$ fine-grained block sparsity allows acceleration in actual modern hardware, it can be used only to accelerate the inference phase. In order to allow for similar accelerations in the training phase, we suggest a novel transposable fine-grained sparsity mask, where the same mask can be used for both forward and backward passes. Our transposable mask guarantees that both the weight matrix and its transpose follow the same sparsity pattern; thus, the matrix multiplication required for passing the error backward can also be accelerated. We formulate the problem of finding the optimal transposable-mask as a minimum-cost flow problem. Additionally, to speed up the minimum-cost flow computation, we also introduce a  fast linear-time approximation that can be used when the masks dynamically change during training. Our experiments suggest a 2x speed-up \revision{in the matrix multiplications} with no accuracy degradation over vision and language models. Finally, to solve the problem of switching between different structure constraints, we suggest a method to convert a pre-trained model with unstructured sparsity to an $N:M$ fine-grained block sparsity model with little to no training.  A reference implementation can be found at \url{https://github.com/papers-submission/structured_transposable_masks}.
\end{abstract}

\section{Introduction}
\label{sec:intro}
Deep neural networks (DNNs) have established themselves as the first-choice tool for a wide range of applications, including computer vision and natural language processing. However, their impressive performance comes at a price of extensive infrastructure costs --- as state-of-the-art DNNs may contain trillions of parameters \citep{Fedus2021SwitchTS} and require thousands of petaflops \citep{gpt3} for the training process. For this reason, compression of DNNs training and inference process is a research topic of paramount importance in both academia and industry. The main techniques of compression include quantization \citep{Banner2018ScalableMF,Nahshan2019LossAP}, knowledge distillation \citep{Hinton2015DistillingTK}, and pruning \citep{Han2015LearningBW,l1Filter}. 

Pruning DNNs is one of the most popular and widely studied methods to improve DNN resource efficiency. The different pruning methods can be categorized into two different groups: unstructured and structured pruning. While the former can achieve a very high compression ratio, it usually fails in reducing the computational footprint in modern hardware. In contrast, structured pruning methods, such as block \citep{blockSparsity} or filter \citep{l1Filter} pruning, are more hardware friendly. Unfortunately, these methods usually fail to keep the original accuracy for high compression ratios \citep{Renda2020ComparingRA}. Finding an optimal structured sparsity pattern is still an ongoing research topic. 

Recently, \citet{Nvidia} announced the A100 GPU, containing sparse tensor cores which are able to accelerate fine-grained sparse matrix multiplication. The sparse tensor cores in A100 enable a 2x acceleration of regular matrix multiplication in DNNs, $Y = WX$, where $W$ and $X$ are weight and input matrices, respectively. The only requirement is that $W$ would have a fine-grained 2:4 sparsity structure, i.e. out of every four contiguous elements in $W$, two are pruned. \citet{Nvidia} suggested a two-fold scheme for pruning a pretrained dense model: (a) Define a fine-grained 2:4 fixed mask, and (b) retrain with the masked weights using original training schedule. Indeed, the \citet{Nvidia} approach is very appealing for the common case where a pretrained dense model is given. 

While the \citet{Nvidia} method works well on many models, a pretrained model is not always given. In those cases, one has to first train a dense model and only then try to prune it. To alleviate this demand, \citet{n:mStructured} suggested a method that trains from scratch a model with $N:M$ fine-grained mask, using a sparse-refined straight-through estimator (SR-STE). Similarly to the quantization-aware-training methods \citep{hubara2017quantized}, they maintain a dense copy of the weights and prune it in every iteration, passing the gradients using the straight-through estimator \citep{bengio2013estimating}. Since the mask dynamically changes while training, they suggest adding an extra weight decay on the masked (i.e. pruned) elements to reduce the mask changes during the training process.
As opposed to \citet{evci2020rigging} that aims to reduce memory footprint for sparse training from scratch, \citet{n:mStructured} only eliminates the need to train a dense model before pruning it.  

\begin{figure*}[h!]
\begin{center}
\includegraphics[width=0.7\columnwidth]{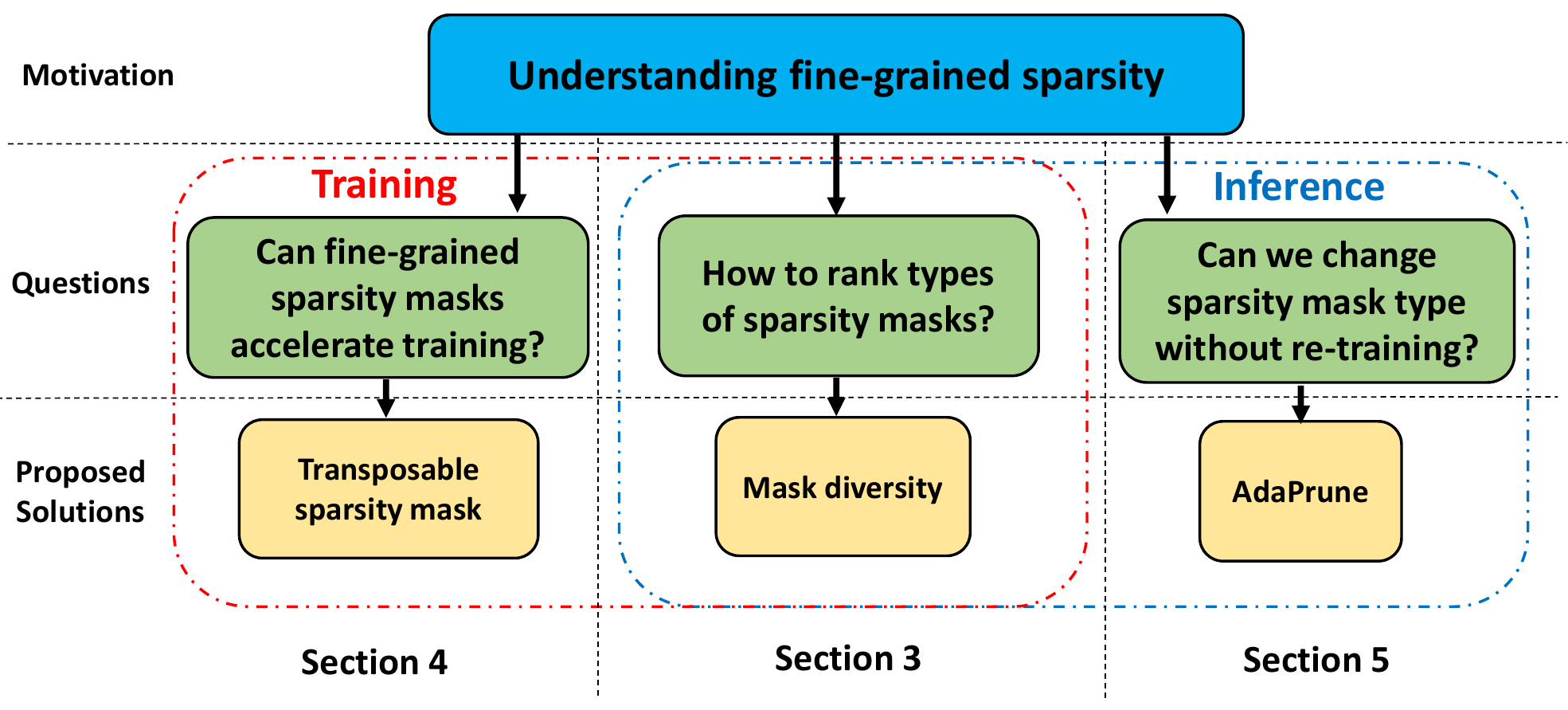}
\caption{\small{High-level overview of the different questions and their corresponding solutions proposed in this work. Motivated by understanding fine-grained sparsity we first suggest a measure to rank different sparsity mask, then we suggest a method to accelerate training with fine-grained sparsity and finally propose a method to change the fine-grained mask without re-training. }}
\label{fig:chart}
\end{center}
\end{figure*}

Motivated by these promising results, our goal here is to answer three remaining questions: (\cref{fig:chart}) 
\begin{enumerate}[wide, leftmargin=*]
    \item \textbf{How to rank different types of sparsity masks?} We suggest a new measure called ``mask diversity", which is the first to connect mask constraints and network accuracy (Section \ref{sec:mask_diversity}).
    \item \textbf{Can fine-grained sparsity masks accelerate training?} We start by observing both the forward and the backward matrix-multiplications involving the weight matrix $W$. Since the backward pass requires using the transposed matrix $W^T$:
    \begin{equation}
    \label{eq:bwd}
    \frac{\partial \mathrm{Loss}}{\partial X} = W^T\cdot \frac{\partial \mathrm{Loss}}{\partial Y}, \
    \end{equation}
    and in general $W^T$ does not have an $N:M$ fine-grained sparsity structure (even if $W$ has this structure),  the methods suggested in \cite{n:mStructured,Nvidia} accelerate only the forward pass matrix-multiplication, $Y=WX$. Consequently, the current methods only utilize in part the sparse tensor cores to accelerate training. We propose a novel $N:M$ transposable-fine-grained sparsity mask, where the same mask can be used for both forward and backward passes (\cref{fig:structuredVsTStructured}). We focus on accelerating sparse training in the two settings detailed above: (a) starting from a pretrained model, and (b) starting from scratch. For (a) we derive a novel algorithm to determine the optimal transposable-mask using a reduction to a min-cost flow problem.  For (b) we devise an approximation algorithm with an (almost) linear (in input-size) time complexity that produces a mask whose $\ell_1$ norm is within a factor of 2 from the optimal mask. We show the effectiveness of both methods (\cref{sec:method}).

\begin{figure*}[h!]
\begin{center}
\begin{subfigure}[b]{0.45\linewidth}
\includegraphics[width=\columnwidth]{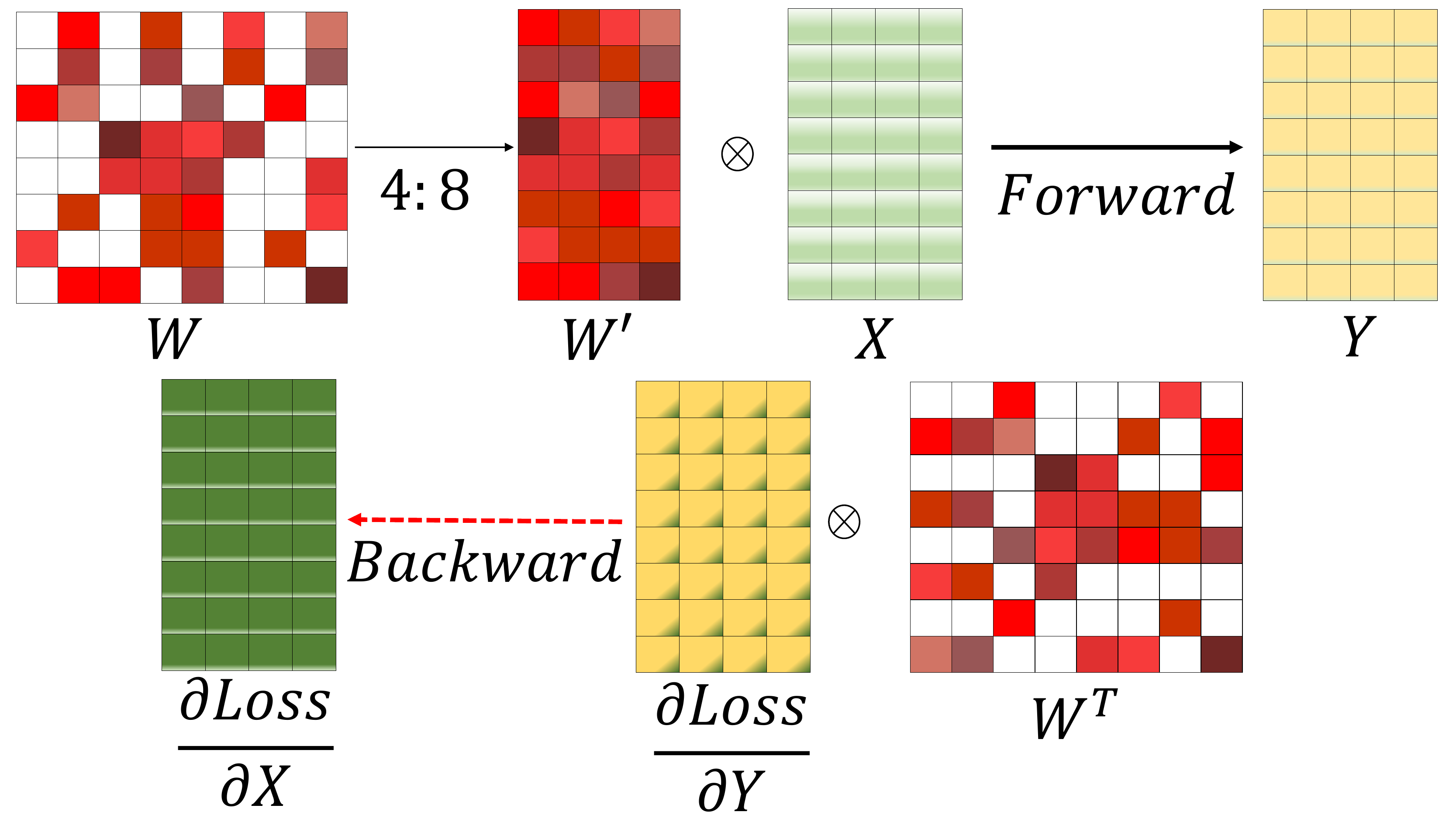}
\caption{}
\end{subfigure}
\hfill
\begin{subfigure}[b]{0.45\linewidth}
\includegraphics[width=\columnwidth]{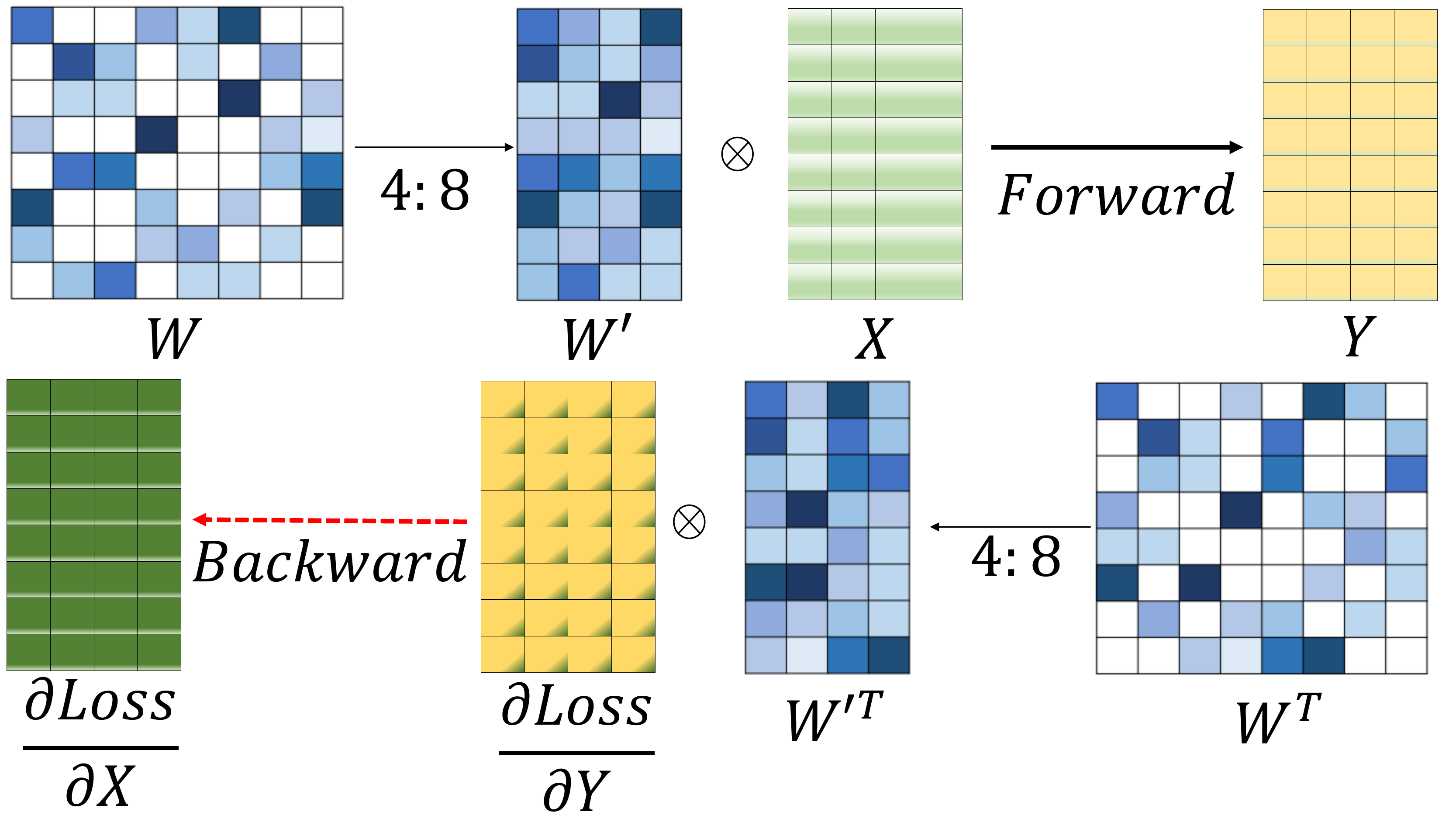}
\caption{}
\end{subfigure}
\caption{\small{\textbf{(a):} A 4:8 structured pruning mask, as  in \citet{n:mStructured,Nvidia}, capable of accelerating with sparse tensors core only the forward pass. \textbf{(b):} The suggested 4:8 transposable structured pruning mask capable of accelerating with sparse tensors core the forward and backward passes.}
}
\label{fig:structuredVsTStructured}
\end{center}\vspace{-4mm}
\end{figure*}

    \item \textbf{Can we change the type of sparsity mask structure without re-training?} Different hardware devices can support different types of fine-grained sparsity masks. Therefore, we suggest the "Adaprune" method, which converts between types of sparsity masks (even from unstructured masks) without the need of re-training, and almost no degradation in accuracy (\cref{sec: adaprune}).
\end{enumerate}

\section{Related work}
\label{sec:related}
Pruning of neural networks weights has been extensively investigated, starting with classical methods in the late 1980s \citep{Janowsky1989,Mozer1989a,Mozer1989b,Karin1990} and then amounting to dozens of papers published in recent years. Since DNNs are generally over-parameterized, pruning the weights reduces their memory footprint. In special cases, when the sparsity mask has a specific pattern, it has the potential to reduce computational footprint as well. The most common practice is to prune a pretrained dense model, so that it will be sparse at deployment. Since the accuracy of the pretrained dense model is known, one can tune the sparsity level to ensure comparable accuracy for the sparse model. Recently, a new line of research that aims to train sparse models from scratch  \citep{gray2017gpu,dettmers2019sparse,evci2020rigging} has emerged. The goal is to train models that cannot fit into currently available hardware. Next, we briefly overview the structured, unstructured, and accelerating-sparse-training categories.

\vspace{-4mm}
\paragraph{Unstructured pruning.} It removes individual elements of the matrix, aiming for high total sparsity, while being agnostic to the location of the pruned elements. Standard pruning methods are based on different criteria, such as magnitude \citep{Han2015LearningBW}, approximate $L_0$ regularization \citep{l0regul}, or connection sensitivity \citep{snip}.  Recent methods \citep{lottery}, suggested to train a dense network until convergence, extract the required mask (``winning ticket"), and use the original training regime to re-train the active weights from their original initialization or final values \citep{rewinding} using the original training schedule. These methods are able to achieve over 80\% sparsity on ResNet50- ImageNet dataset \citep{rewinding}.  Despite the high sparsity ratio that can be achieved with these methods, modern hardware cannot efficiently utilize such a form of sparsity for reducing computational resources \citep{Nvidia}. 
\vspace{-4mm}
\paragraph{Structured pruning.} It removes weights in specific location based patterns, which are more useful for hardware acceleration. Such methods can be applied at either the level of channels or layers. For example, \citet{l1Filter} remove the channels with the lower norm, \citet{thinet} prune channels according to the effect on the activation of the following layer, and \cite{blockSparsity} split the filters
into multiple groups, applying a group Lasso regularization. All these methods are natively supported in both hardware and software, as they effectively change the model structure by reducing channels or groups. Yet, no such method was able to achieve a reasonable accuracy with sparsity levels higher than 50\%. As observed by  \citet{liu2018rethinking}, filter pruning of a pretrained dense over-parameterized
model is rarely the best method to obtain an efficient final model. Thus, here the structured pruning serves mostly as a DNN architecture search for the optimal compact model \cite{tan2019mnasnet,wu2019fbnet}. 
Our work is most closely related to \citet{n:mStructured}, which is the first work that attempted training with a fine-grained $N:M$ structured sparsity mask, as explained above.
\vspace{-4mm}
\paragraph{Accelerating sparse training for model deployment.}
The most common approach for sparse model deployment requires a three steps process: (a) train a dense model; (b) define a sparsity mask (c) fine-tune while enforcing the mask on the model's weights. The lottery ticket hypothesis of \citet{lottery} demonstrates that we can avoid the dense training, i.e., step (a), had we known how to choose the appropriate mask. Since \citet{lottery} discovered the optimal mask (winning ticket) by applying dense training, the question of how to find the optimal mask without training remained open. Since setting a predefined fixed  mask results in some accuracy degradation (\citet{gray2017gpu} that requires expanding the model size), several researchers \citep{bellec2017deep,mocanu2018scalable,mostafa2019parameter,dettmers2019sparse,evci2020rigging} tried to enable dynamic mask changes during the training process. All methods focused on unstructured sparsity and aimed to enable training large models on hardware with memory limitations.
In contrast to these approaches, \citet{n:mStructured} does not aim to reduce the model memory footprint during training, but rather accelerate inference while avoiding dense pre-training. To that end, they keep a dense weight matrix, and in each iteration they re-calculate the mask. With the obtained pruned copy of the weights they perform the forward and backward pass and update the dense copy. Therefore, this method is most relevant when a pretrained dense model is not given, and one wishes to obtain a fine grained $N:M$ sparse model for inference. While our work focuses on the setting of \citet{n:mStructured}, we argue that it can easily combined with  \citet{evci2020rigging} work, as it aims to solve a different issue within the same problem.


\section{Mask Diversity}
\label{sec:mask_diversity}
Structured sparsity requires masks with a hardware-friendly structure type. Yet, which structure should be employed is still an open question. The additional hardware cost (mainly chip-area and power) required to support each of the structures is hard to quantify, as it varies based on the hardware design. However, the effect of the structure on the model accuracy is oblivious to the hardware at hand. Therefore, in this section, we aim to find a method for ranking different types of sparsity masks, which can predict help the model accuracy. We start from an hypothesis that the structure constraints cause accuracy degradation. Thus, we expect that the best sparsity levels, without accuracy degradation, would be achieved by unstructured sparsity, which has no requirements on the sparsity structure type. To quantify how much a specific structure constrains the model, we introduce a new measure, called \textit{mask-diversity} (MD). MD is the number of all possible masks that adhere to the  restrictions of the structure under similar sparsity level. As an example, we derive the MD for a tensor size $T$ under four different structure constraints: 
\begin{enumerate}[wide, leftmargin=*]
    \item \textbf{Unstructured sparsity:} no constraints, except for an overall sparsity level. This is the most common setting in the literature. 
    \item \textbf{Structured N:M sparsity:}  $N$ values in each block of size $M$ are set to zero. This structure is currently supported in the most widely deployed hardware  \citep{Nvidia}. 
    \item \textbf{Transposable N:M  sparsity:} for a block of size $M\times M$ both columns and rows must follow the $N:M$ fine-grained constraints. As we later discuss (\cref{sec:method}), the transposable structure is essential for training acceleration.    
    \item\textbf{Sequential N:M sparsity:} any  block of size $M$ must contain $N$ sequential zeros. This is an example for a small potential modification to the $N:M$ fine-grained structure, which might be more hardware friendly (data can be compressed and decompressed more easily).
\end{enumerate}    
MD depends on the required sparsity level (and the tensor size), thus without loss of generality we set the sparsity level to be $N/M$. In \cref{eq:mask_d} we write the MD for each of the constraints (1-4) above, derived using basic combinatorial arguments (\cref{app:MD_deriv}):
\vspace{-1mm}
\begin{align}
\begin{split}
 &\textrm{\textbf{1.}}\hspace{3pt} \mathrm{MD}_{\textrm{\shortstack{Unstructured}}}=\binom{T}{\frac{NT}{M}}; \hspace{25pt} \textrm{\textbf{3.}}\hspace{3pt} \mathrm{MD}_{\textrm{Transposable}} =\left(M!\left(M-1\right)!\cdot\cdot\cdot\left(M-N+1\right)!\right)^{\frac{T}{M^2}}\\
 & \textrm{\textbf{2.}}\hspace{3pt} \mathrm{MD}_{\textrm{\shortstack{Structured}}} = \left(\frac{M!}{N!\left(M-N\right)!}\right)^{\frac{T}{M}}; \hspace{10pt} \textrm{\textbf{4.}}\hspace{3pt}  \mathrm{MD}_{\textrm{\shortstack{Sequential}}} =(M-N+1)^{\frac{T}{M}}
 \end{split}
 \label{eq:mask_d}
\end{align}

\begin{wraptable}{r}{8cm}
\vspace{-4mm}
\centering
\caption{\small{MD for different constraints for a matrix of size $8 \times 8$.}}
\label{tab:diversity}
\begin{tabular}{l|c|c|c}
\toprule
$N:M$ & 1:2 & 2:4 & 4:8 \\ \hline
Unstructured & $1.8\cdot10^{18}$ & $1.8\cdot10^{18}$ & $1.8\cdot10^{18}$ \\ \hline
Structured & $4\cdot10^9$ & $2.8\cdot10^{12}$ & $5.7\cdot10^{14}$ \\ \hline
Transposable & $6\cdot10^4$ & $4\cdot10^8$ & $1.7\cdot10^{13}$ \\  \hline
Sequential & $4\cdot10^9$ & $4\cdot10^7$ & $4\cdot10^5$  \\\bottomrule
\end{tabular}
\vspace{-2mm}
\end{wraptable}

\begin{figure*}[h!]
\vspace{-2mm}
\begin{center}
\begin{subfigure}[b]{0.5\linewidth}
\includegraphics[width=\columnwidth]{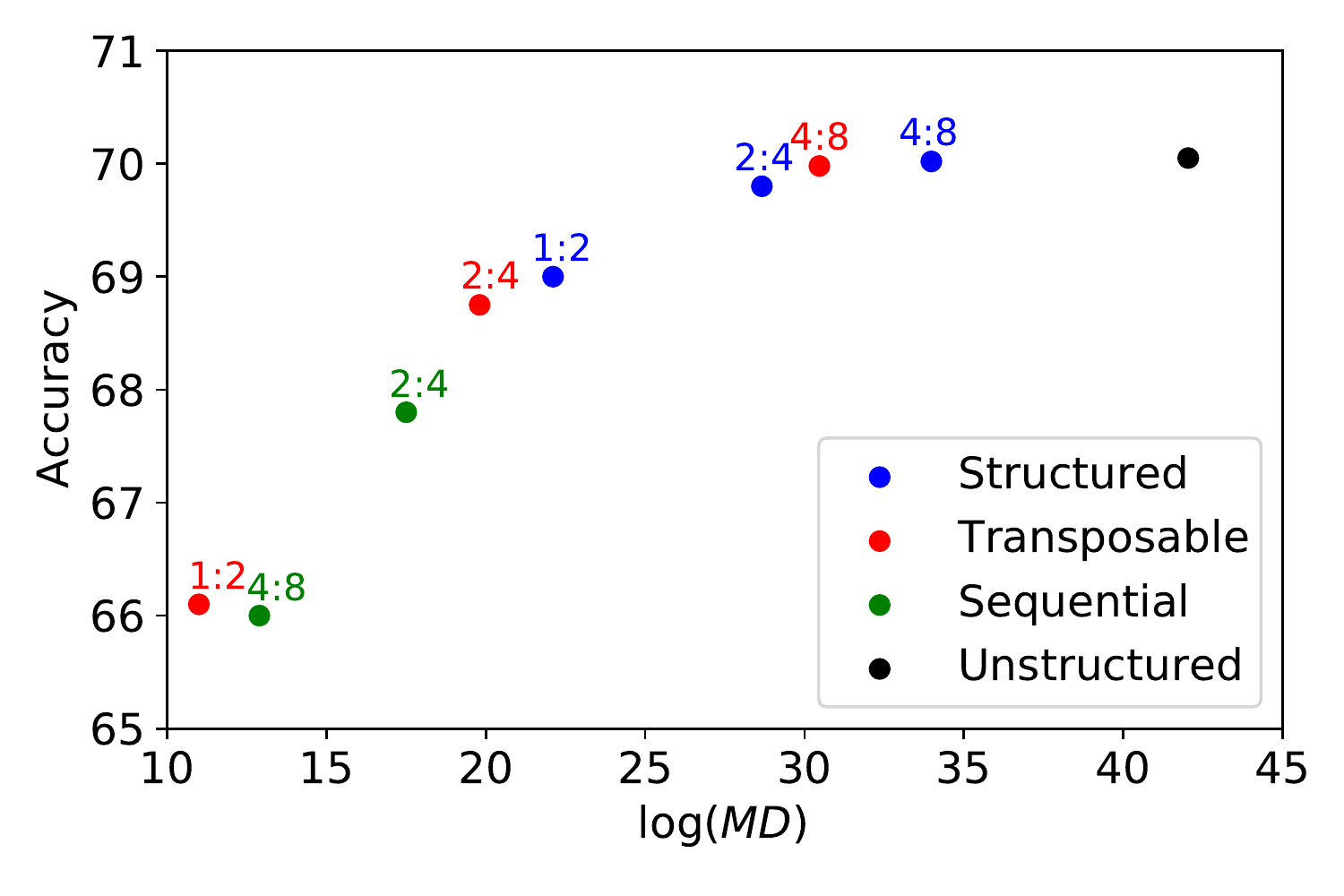}
\caption{}
 \label{fig:diversity}
\end{subfigure}
\begin{subfigure}[b]{0.49\linewidth}
\includegraphics[width=\columnwidth]{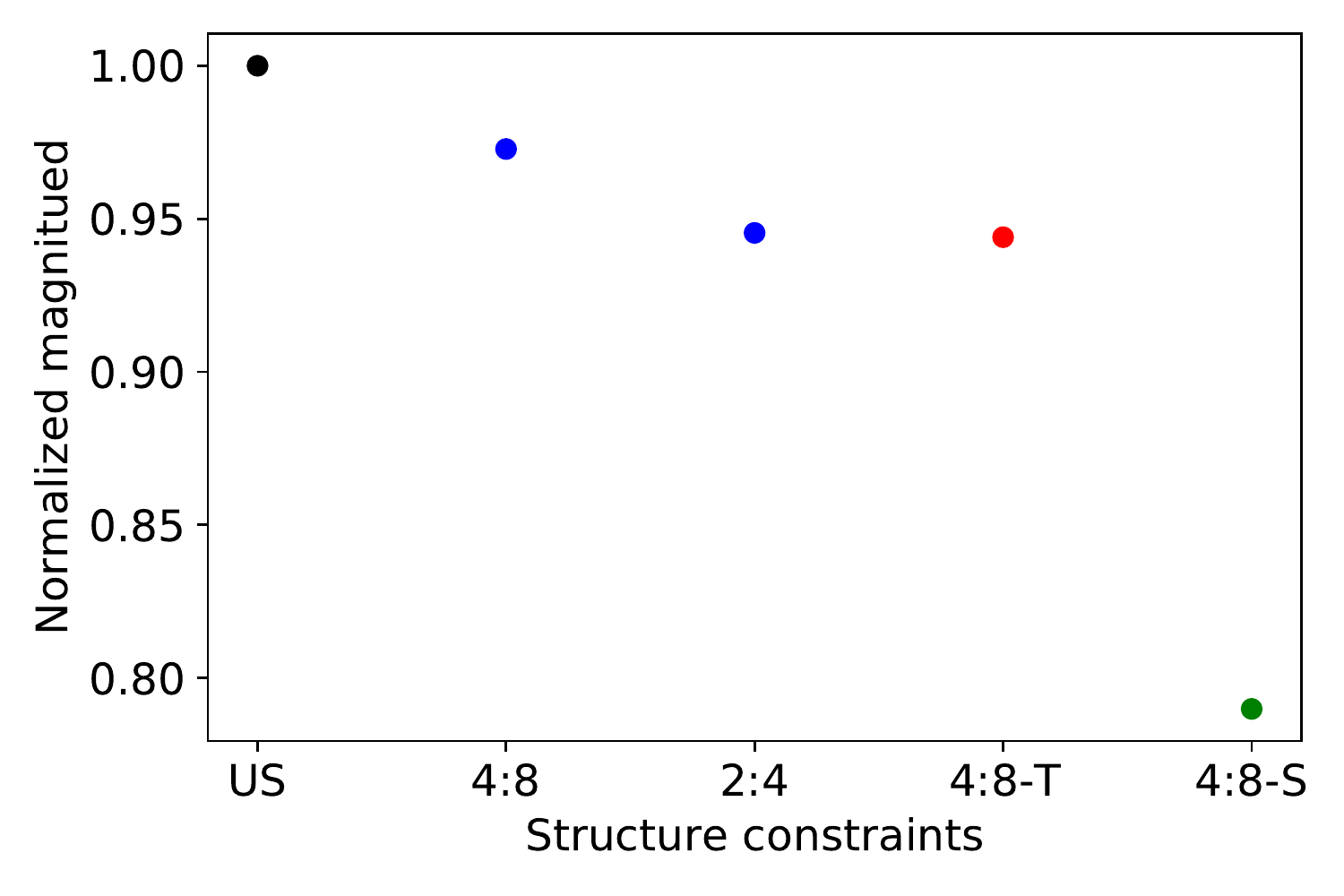}
\caption{}
\label{fig:magnitude_8_8}
\end{subfigure}
\caption{\small{\textbf{(a): }ResNet18 over Cifar100 top-1 accuracy for weight sparsity of 50 \% using different structured and unstructured masks. As expected the mask diversity correlates with the pruned model accuracy. \textbf{(b):} Magnitude of the last layer's weight tensor of ResNet-50 (pretrained dense model) masked with structured mask 4:8, 2:4, 4:8 transposable ("4:8-T") and 4:8 sequential ("4:8-S") normalized by the unstructured  50\% sparsity ("US"). Notice that mask diversity is correlated with magnitude preservation. As expected the 4:8 transposable mask has a similar $\ell_1$ norm score as the 2:4 mask. Additional results and details in \cref{app:MD_exp}  }.}

\end{center}
\end{figure*}

\cref{tab:diversity} shows the MD for a matrix of size $8 \times 8$ with 50\% sparsity under different constraints. As expected, unstructured sparsity achieves best in class MD, however its hardware cost makes it unfeasible to use on modern architectures \citep{gray2017gpu,Elsen2020FastSC,Gale2020SparseGK}. Additionally, sequential N:M sparsity has worst in class MD and consequently we expect that using it would harm the model accuracy (\cref{fig:diversity}). Thus, from here on, we focus on the structured and transposable structured fine-grained sparsity. Notice the MD of structured 2:4, which was used in \citep{Nvidia,n:mStructured} is severely reduced  when transposable constraints are enforced. Since a transposable structure enables training acceleration (\cref{sec:method}), supporting this structure is essential. Thus, we suggest a minimal expansion of the block size (from 4 to 8) which enables the transposable fine-grained structure to reach a similar MD as the 2:4. Increasing the block size requires more multiplexers to support the $N:M$ structure \citep{liu2020sparse}. However, the hardware cost increases only logarithmically with respect to $N$ (details given in \cref{sec:appHW}). We therefore argue that such an expansion is reasonable. 
To better test our hypothesis, we measured the accuracy degradation with respect to the mask diversity on ResNet18 with Cifar-100 dataset. As can be seen in  \cref{fig:diversity}, the accuracy correlates with the MD metric and the 4:8 transposable fine-grained mask reached a comparable accuracy to the 2:4 fine-grained mask. Moreover, recall the common method of magnitude pruning has an underlying objective to preserve the tensor $\ell_1$ norm. We therefore measured the effect of the different constraints on the $\ell_1$ norm in \cref{fig:magnitude_8_8}. This figure shows the $\ell_1$ norm of the last layer in trained ResNet-50 after applying different structured pruning. Here as well, 2:4 structured and 4:8 transposable structured have (almost)  similar $\ell_1$ norms. 

\vspace{-6mm}

\section{Computing transposable sparsity masks}
\label{sec:method}

In general, training DNNs requires three matrix multiplications per layer. The first multiplication is required for the forward propagation between the weights and activation. The other two multiplications are used for the backward and update phases. The backward phase calculates the gradients of the loss function with respect to the input of the neural layer. This is done by recursively passing the error from the last layer to the first (\cref{eq:bwd}). Note that the backward phase uses the transposed weight matrix. Hence, accelerating the backward phase requires the transposed weight matrix to adhere to the hardware required pattern (e.g., $N:M$ fine-grained sparsity). In this section, we tackle this issue by presenting a novel to find $N:M$ transposable fine-grained sparsity masks, where the same mask can be used to accelerate both forward and backward passes (\cref{fig:structuredVsTStructured}). The required mask contains only $M-N$ non-zero elements, for every contiguous $M$ elements, in both $W$ and $W^T$ simultaneously. We formulate the problem and suggest two methods to generate the transposable mask.


\textbf{Problem formulation.}
First, we provide an integer-programming (IP) formulation for finding an optimal transposable fine-grained mask. Let us consider a block of size $M \times M$ in a  weight matrix $W$. Our goal is to maximize the $\ell_1$ norm of  $W$ after masking N elements in each row and column. We formulate the problem as an integer program. Define a binary sparsity mask $S\in\left\{ 0,1\right\} ^{M\times M}$, where $S_{i,j}=1$ if and only if the element $W_{i,j}$ is not pruned, and otherwise $S_{i,j}=0$. The resulting integer program is
\begin{equation}
\label{eq:ip}
\max_{{S}\in\left\{ 0,1\right\} ^{M\times M}}\sum_{i,j}S_{i,j}\left|W_{i,j}\right|\,\,\mathrm{s.t.}\,\,\forall j:\sum_{i}S_{i,j}=N\,,\forall i:\sum_{j}S_{i,j}=N\,.
\end{equation}
In the following, we examine several methods for solving this problem. We first describe an optimal, yet computationally expensive method. We then describe a more efficient method, which provides an approximate near-optimal solution.

\begin{figure}[h!]
\begin{center}
\centerline{\includegraphics[width=0.8\columnwidth]{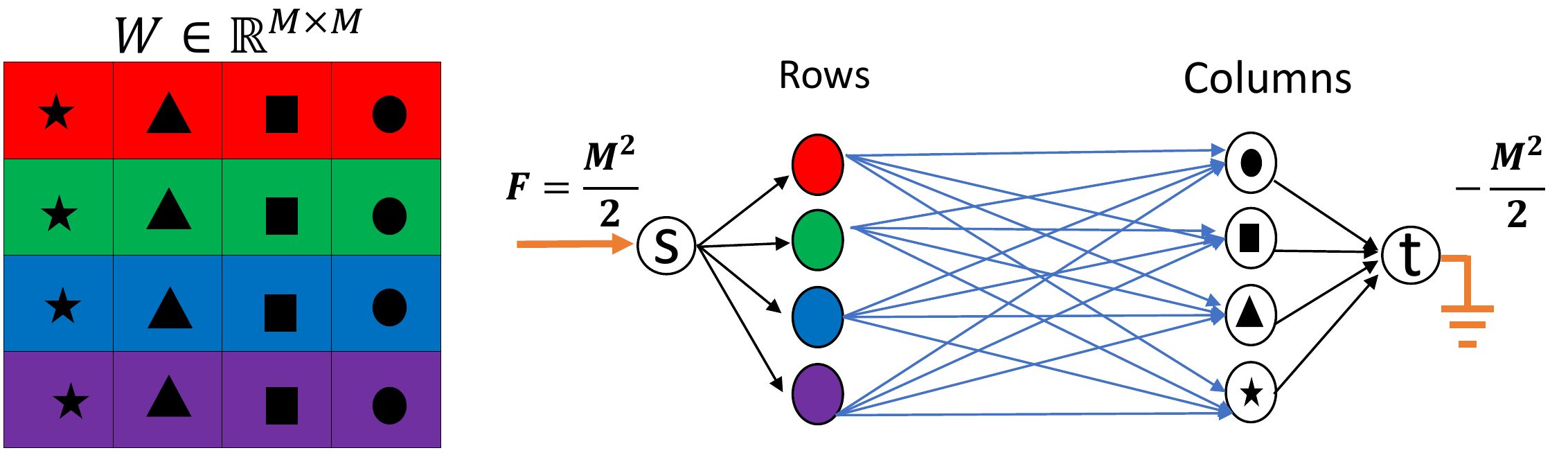}}
\caption{\small{$\frac{M}{2}:M$ transposable-sparsity optimization as a min-cost flow problem. In addition to a source and a sink, the network has a node for each row and for each column. The construction uses three types of edges: (i) \textit{source edges} emanating from the source node $s$ into each row node $i$; (ii) \textit {sink edges} connecting each column node $j$ with the  sink node $t$; and (iii) a \textit{coefficient edge} $(i,j)$ for each matrix element $ W_{i,j}$. Each source edge $(s,i)$ has capacity $\frac{M}{2}$ which is equal to the number of elements that need to be selected for pruning in row $i$. Similarly, each sink edge $(j,t)$ has capacity $\frac{M}{2}$ which is equal to the number of elements pruned in column $j$. Each coefficient edge $(i,j)$ has unit capacity and cost $\left| W_{i,j}\right|$. Finally, selecting a matrix element with weight $ W_{i,j}$ for pruning corresponds to a unit flow on the coefficient edge $(i,j)$.  Assuming the source and sink edges have zero-cost, there is a one-to-one correspondence between a min-cost flow solution that sends a flow of value $\frac{M^2}{2}$ from the source $s$ to the destination $t$ in this construction, and an optimal transposable mask minimizing the sum of absolute values  selected for pruning. }  }
 \label{fig:minCost}
\end{center}
\vskip -0.2in
\end{figure}

\textbf{Reduction to Min-Cost Flow.}
General integer programs (IP) require exponential time complexity with respect to the input size (worst-case). Fortunately,  \cref{fig:minCost} shows that our IP formulation (\cref{eq:ip}) reduces to a min-cost flow problem. There is a vast literature on efficient min-cost flow algorithms (see e.g., \citep{ahuja1988network}), however, the most efficient algorithms for our setting take time $O(M^3\log M)$ time for computing an optimal transposable mask for a block size of $M\times M$ \citep{ahuja1988network}[pp. 396-397]\footnote{More modern methods that are based on interior point algorithms seem to be less efficient for our setting.}. The min-cost flow solution should be used when training from a pretrained dense model, where the transposable mask is generated once, remaining fixed from then on during training. On the other hand, sparse training from scratch requires changing the mask during training, and it is therefore essential to find a very efficient algorithm for computing the mask. To this end,  we design a light \textit{2-approximation} algorithm, i.e., for every input it produces a solution which is guaranteed to be within a factor of 2 of an optimal solution (to the given input), yet it runs in almost linear time.
\paragraph{2-Approximation algorithm.}
\label{sec:approxi}
We design a greedy \textit{2-approximation} algorithm (see \cref{alg:approx}) having a low time complexity that can be used in practice without compromising too much the quality of the solution produced. Unlike the optimal min cost flow solution that runs in time complexity of $O(M^3)$ for a block size of $M\times M$,  \cref{alg:approx} has a running time of $O(M^2\log M)$ i.e., a time complexity that is almost linear in the number of block elements $M^2$. The approximation algorithm uses the same construction described in \cref{fig:minCost}, but instead of running a min-cost flow on the graph, it employs a simple greedy approach. Let $P$ be the list of edges pruned by \cref{alg:approx}, let $W(P)$ be the total weight of the edges in $P$, and  let $W^{*}$ be the weight of an optimal solution (i.e., the minimal sum of edges that can be pruned to create a $\frac{M}{2}:M$ transposable sparsity mask). The next lemma establishes that  \cref{alg:approx} finds a 2-approximate solution (proof in \cref{app:lema proof}, with an example showing the upper bound is tight):
\begin{theorem}
Algorithm 1 produces a tight 2-approximate solution, i.e., $W(P)<2\cdot W^{*}$.
\end{theorem}
\begin{algorithm}
  \caption{2-approximation algorithm. {\bfseries Input:} Bipartite graph G=(V,E) ;  {\bfseries Initialize:} P= $\emptyset$} 
  \label{alg:approx}
  \begin{algorithmic}[1]
    \State Sort list of coefficient edges from light to heavy; Let $A = [ e_1,...,e_n ]$ be the sorted list of edges.
    \For{each edge $e_i = (u,v)\in A$, $i=1, \ldots,n$}
    \If{degree(u)$\leq \frac{M}{2}$ or degree(v)$\leq \frac{M}{2}$ in P} 
    $\text{P} \gets \text{P} + e_i$.
    \EndIf
    \EndFor
\end{algorithmic}
\end{algorithm}

In Table \ref{running_resnet50} we show the running time overhead of ResNet50 training with IP, min cost flow and 2-approximation algorithms over regular training, the algorithms were implemented in a non-optimized way. All experiments were run in a single GPU and the mask was updated every 40 iterations. Notice the acceleration achieved with the 2-approximation algorithm in comparison to the naive IP.  In \cref{sec:appRunTime} we extend the complexity analysis.

\begin{wraptable}{r}{8cm}
\vspace{-0.4cm}
\centering
\caption{\small{Overhead of different algorithms for finding the $4:8$ transposable mask: ratio of their running time over regular training (ResNet50). Notice the overhead reduction in the 2-approximation algorithm in comparison to the naive IP.}}
\label{running_resnet50}
\begin{tabular}{l|c}
\toprule
 Method &  Overhead (\%)   \\ \hline
Integer-programming  & 180 \\ \hline
Min-cost flow  &  70 \\ \hline
2-approximation & 14 \\ \bottomrule
\end{tabular}
\vspace{-0.3cm}
\end{wraptable}

\subsection{Experiments}
\label{sec:experiments}
In this section, we demonstrate the effectiveness of our proposed transposable $N:M$ fine-grained structured sparsity in computer vision and natural language processing tasks. We evaluate the suggested method in two cases: (i) initialize from a trained dense model and re-train with a fixed mask, similar to APEX’s Automatic Sparsity (ASP \cite{Nvidia}),  (ii) train from scratch and update the mask frequently, as done by \citet{n:mStructured}.
We show comparable accuracy to previous methods, while achieving a significant reduction of the training process resources --- by exploiting the sparse tensor core abilities, allowing their use both in forward and backward passes. In all the experiments we use a  4:8 transposable-mask, which as shown in \cref{tab:diversity}, have a similar MD as 2:4 mask used in previous works \citep{Nvidia,n:mStructured}. Experimental details appear in \cref{sec:appExpSettings}. \revision{In \cref{sec:AppExp24} we show additional experiments with the 2:4 transpose mask showing that: (1) in most cases,  2:4 transpose is enough to achieve high accuracy, (2) in some cases where the 4:8 transpose is necessary, and (3) this is consistent with the MD results shown in \cref{sec:mask_diversity}}.

\textbf{Initialization from a trained dense model}
We evaluate the suggested $N:M$ transposable mask using a trained dense model as initialization. In order to find the transposable mask, we solve the min-cost flow reduction (\cref{fig:minCost}) on the dense trained network and then fix the mask.
In \cref{tab:2stages} we compare our method with ASP \citep{Nvidia} on classification (ResNet50 - ImageNet dataset), detection (MaskRCNN - COCO dataset) and question answering (BERT-large - SQuAD dataset) tasks.
While both methods initialized from the pre-trained models, the 4:8 transposable enables propagation acceleration in the retraining phase where ASP does not.

\vspace{-0.3cm}
\begin{table}[h]
\centering
\caption{\small{Comparison of our suggested method with ASP \citep{Nvidia} initialized from dense model on ResNet50 (Imagenet dataset), BERT-large (SQuAD dataset) and MaskRCNN (COCO dataset). We use a transposable 4:8 mask while ASP used 2:4. SU refers to the sparse tensor cores utilization, used both in forward and backward passes in the proposed method, allowing 2x \revision{sparse tensor cores utilization} in comparison to ASP which use the sparse tensor cores only in the forward pass.  Missing results in \citet{Nvidia} were marked by '-'}}
\label{tab:2stages}
\begin{tabular}{l|l|c|c|c|c|c|c}
\toprule
\multirow{2}{*}{Model} & \multirow{2}{*}{Metric} & \multicolumn{2}{c|}{Baseline}  & \multicolumn{2}{c|}{ASP\citep{Nvidia} 2:4} & \multicolumn{2}{c}{Ours 4:8-T}                                              \\ \cline{3-8}  & & \multicolumn{1}{c|}{SU} & \multicolumn{1}{c|}{Accuracy} & \multicolumn{1}{c|}{SU} & \multicolumn{1}{c|}{Accuracy} & \multicolumn{1}{c|}{SU} & \multicolumn{1}{c}{Accuracy} \\ \hline
ResNet18   & Top1  & \multirow{4}{*}{0\%}   & 69.7\%    & \multirow{4}{*}{33\%}  & -   & \multirow{4}{*}{66\%}   & 70.06\%  \\ \cline{1-2} \cline{4-4} \cline{6-6} \cline{8-8} 
ResNet50   & Top1   &  & 76.15\%   & & 76.6\%  & & 76.6\%                \\ \cline{1-2} \cline{4-4} \cline{6-6} \cline{8-8} 
Bert-Large    & F1  && 91.1 &  & 91.5 & & 91.67                         \\ \cline{1-2} \cline{4-4} \cline{6-6} \cline{8-8} 
MaskRCNN & AP &  & 37.7 &  & 37.9 & & 37.84   \\ \bottomrule
\end{tabular}
\end{table}

\textbf{Sparse Training from scratch.}
In order to avoid the training of a dense model, we also evaluate the proposed transposable $N:M$ mask in  the training from scratch setting. Similar to \citet{n:mStructured} we keep a dense copy of the weights and before each forward pass we mask the weights with a $N:M$ transposable mask. In contrast to \citet{n:mStructured} who changed the mask every iteration, we found that we can use 2-approximation scheme to extract the transposable mask every 40 iterations. Empirically we found that the 2-approximation scheme is on average within a factor of 1.2 from the optimal mask. The hyper-parameters used for training are equal to the ones suggested by \citet{n:mStructured}. In \cref{tab:fromScratch} we test the proposed method over ResNet18, ResNet50, ResNext50, Vgg11 (ImageNet dataset) and fine-tune of Bert (SQuAD-v1.1 dataset) and compare to \citet{n:mStructured} results. As can be seen, we achieved comparable accuracy with 2x \revision{sparse tensor cores utilization} in the training process.

\begin{table}[]
\centering
\caption{\small{Training from scratch of ResNet18, ResNet50, ResNext50 and Vgg11 on imageNet dataset and fine-tuning of Bert-base on SQuAD dataset,  using the proposed 2-approximation scheme. SU refers to the sparse tensor cores utilization, used both in forward and backward passes in the proposed method, allowing 2x \revision{sparse tensor cores utilization} in comparison to N:M-SS \citep{n:mStructured} with comparable results. Missing results in \citet{n:mStructured} were marked by '-'}}
\label{tab:fromScratch}
\begin{tabular}{l|l|c|c|c|c|c|c}
\toprule
\multirow{2}{*}{Model} & \multirow{2}{*}{Metric} & \multicolumn{2}{c|}{Baseline} & \multicolumn{2}{c|}{4:8-SS \citep{n:mStructured}} & \multicolumn{2}{c}{Ours 4:8-T} \\ \cline{3-8} 
 &  & \multicolumn{1}{c|}{SU} & \multicolumn{1}{c|}{Accuracy} & \multicolumn{1}{c|}{SU} & \multicolumn{1}{c|}{Accuracy} & \multicolumn{1}{c|}{SU} & \multicolumn{1}{c}{Accuracy} \\ \hline
ResNet18 & Top1 & \multirow{5}{*}{0\%} & 70.54\% & \multirow{5}{*}{33\%} & 71.2\% & \multirow{5}{*}{66\%} & 70.75\% \\ \cline{1-2} \cline{4-4} \cline{6-6} \cline{8-8} 
ResNet50 & Top1 &  & 77.3\% &  & 77.4\% &  & 77.1\% \\ \cline{1-2} \cline{4-4} \cline{6-6} \cline{8-8} 
ResNext50 & Top1 &  & 77.6\% &  & - &  & 77.4\% \\ \cline{1-2} \cline{4-4} \cline{6-6} \cline{8-8} 
Vgg11 & Top1 &  & 69\% &  & - &  & 68.8\% \\ \cline{1-2} \cline{4-4} \cline{6-6} \cline{8-8} 
BERT-base & F1 &  & 88.52 &  & - &  & 88.38 \\ \bottomrule
\end{tabular}
\end{table}

\section{Structured sparsity without full training  \label{sec: adaprune}}
\begin{figure*}[h!]
\begin{center}
\begin{subfigure}[b]{0.46\linewidth}
\includegraphics[width=\columnwidth]{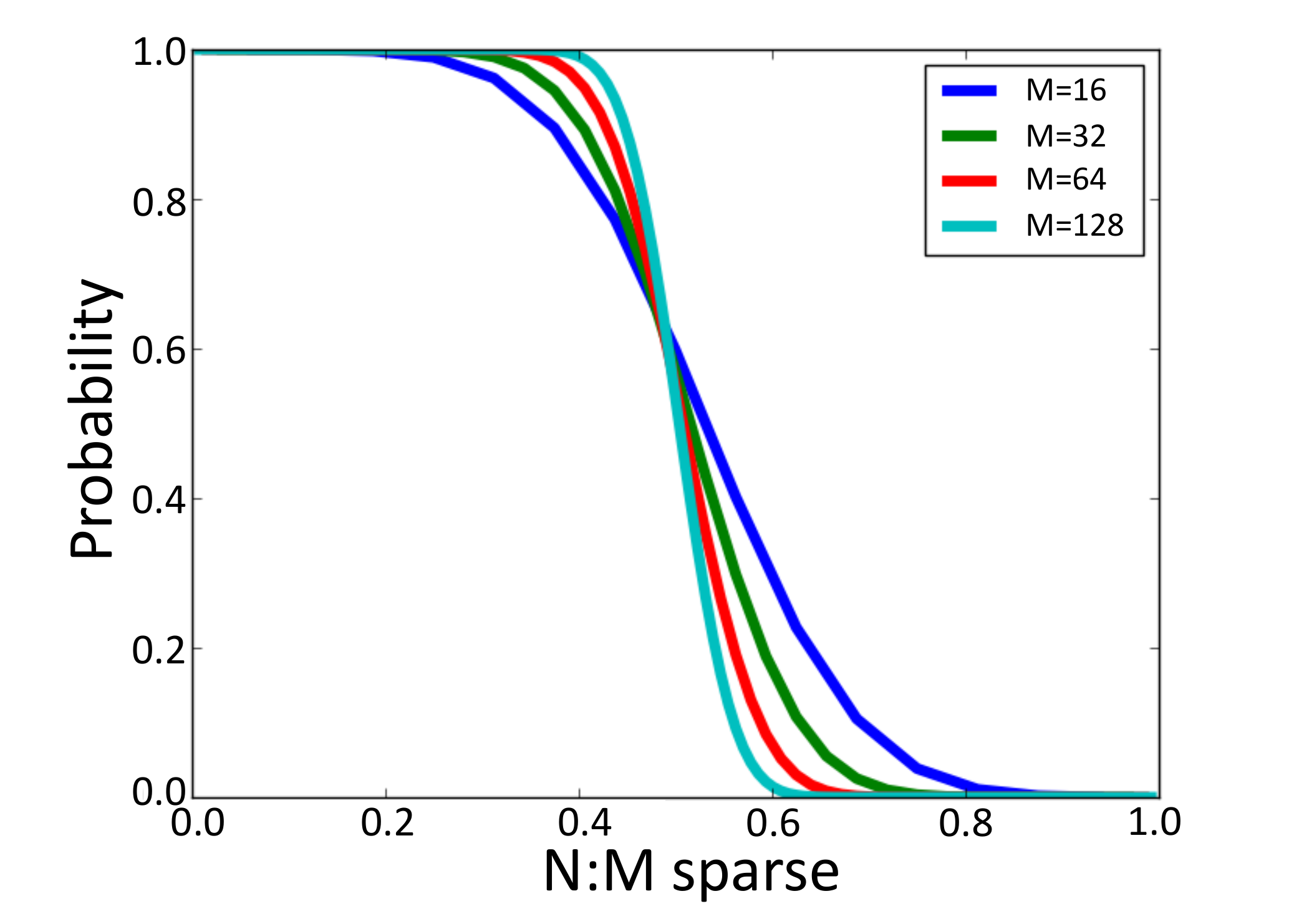}
\vspace{-1.5\baselineskip}
\caption{}
 \label{fig:coffee}
\end{subfigure}
\hfill
\begin{subfigure}[b]{0.49\linewidth}
\includegraphics[width=\columnwidth]{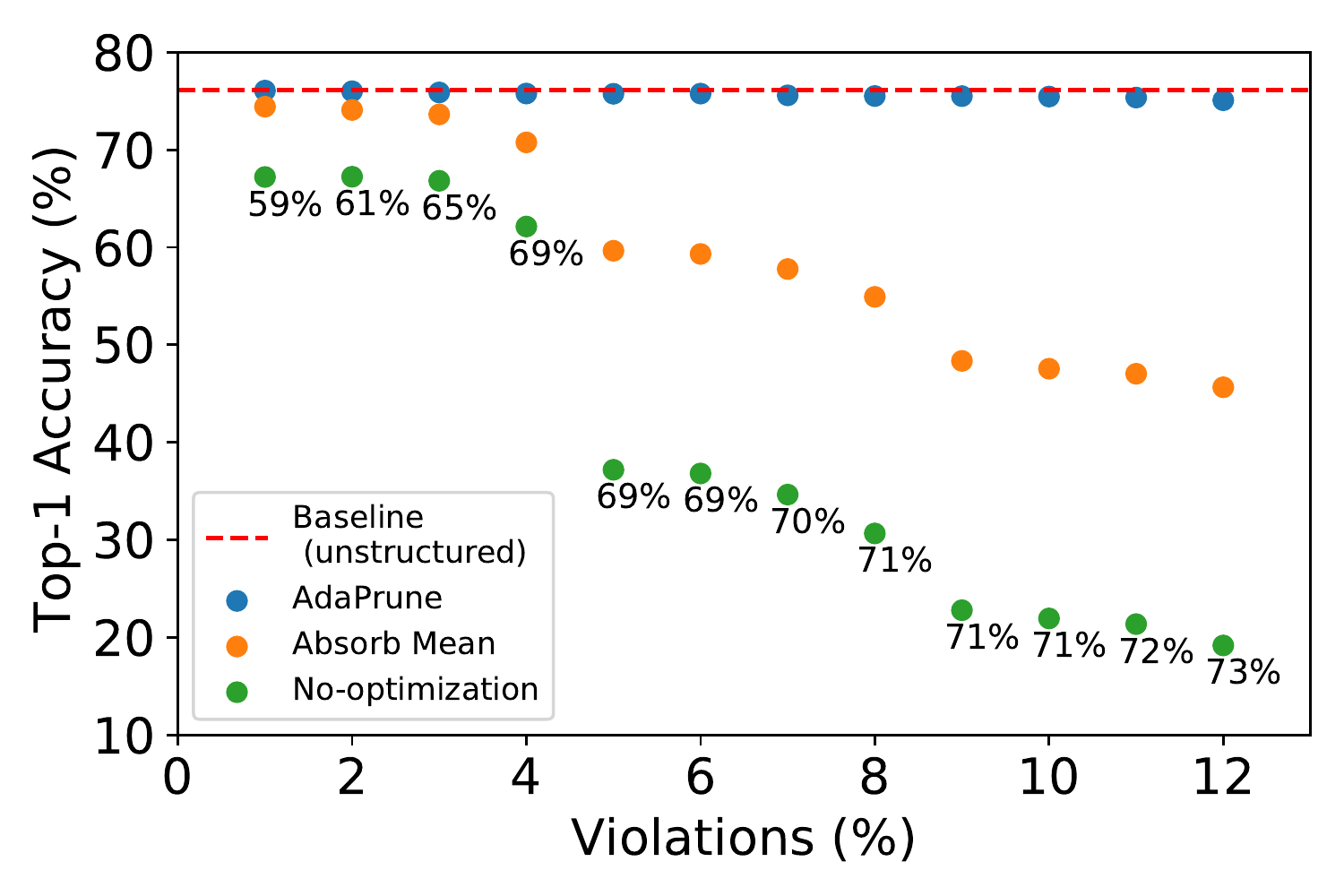}
\vspace{-1.5\baselineskip}
\caption{}
 \label{fig:violation_mean_adaprune}
\end{subfigure}
\vspace{-0.4\baselineskip}
\caption{\small{\textbf{(a):} \cref{eq:P_sparse} for $\rho=0.5$ and various block sizes $M$. We have a sharp ("phase") transition at $N/M = \rho$. Specifically, (i) when $N/M \leq \rho$ we have a probability larger than 0.5 that the sampled block is $N:M$ sparse; (ii)  when $N/M \geq \rho$ this probability quickly decreases to zero.  As block size $M$ increases this phase-transition gets sharper. As expected, when $M \to \infty$, unstructured sparsity satisfies the structured constraints, and we expect it to display the phase transition precisely at the critical point $\rho$. \textbf{(b): }Top-1 accuracy vs. percent of constraints violated. The numbers next to the baseline samples represents the sparsity level of the refined model. }}
\end{center}
\vspace{-6mm}
\end{figure*}

In section \cref{sec:method} we suggested 4:8 transposable-mask to accelerate training, but what should we do if we wish to deploy the output of such training on hardware that supports only 2:4 fine-grained sparsity. Forcing structured sparsity on a model that was trained with a different structured sparsity, leads to a severe accuracy degradation as several bits of the mask may change to satisfy the structured sparsity requirements. In this section we would focus on the more common case of deploying unstructured sparse model on hardware that support N:M structure.
This is a fundamental problem as most DNN pruning methods focus on unstructured pruning, which reduces the memory footprint. However, current hardware implementations suggest that, unless very high sparsity levels are achieved, the model cannot be accelerated at all. Hence, commonly, the weights are simply decompressed before multiplication. To understand the problem 
we study the probability that an unstructured mask would not violate any $N:M$ constraint \citep{Nvidia}. Then we discuss two light methods to bridge the gap when a sparse model is given but the hardware does not support its structure.
\vspace{-2mm}
\paragraph{Probability for violating the $N:M$ constraint in unstructured sparsity.} Let $X=\{x_1, x_2, ..., x_M\}$ be a block of independent and identically distributed random variables. Assume that with a probability $\rho$, $x_i$ can be pruned without accuracy degradation (i.e., unstructured pruning). In this section, we consider a general form of block sparsity in which, for a block of size $M$, at least $N$ values could be pruned. Define $X$ to be $N:M$ sparse if this $M$ sized block has at least $N$ values that can be pruned without harming accuracy. The probability of having a $N:M$ sparse block is given by the binomial distribution and so
\begin{align}
      P\left( X \text{ is } N:M \text{ sparse}\right) 
      = \sum_{i\geq N}  \binom{M}{i} \cdot \rho^i \cdot (1-\rho)^{M-i}\,
     \label{eq:P_sparse}
\end{align}
In \cref{fig:coffee} we plot \cref{eq:P_sparse} for  the case of $\rho=0.5$. To force a given sparse model to have a fine grained $N:M$ sparsity, we need to make sure that  $N$ out of every $M$ contiguous elements are zero. Therefore, as in \citet{Nvidia}, in each block we prune $N$ weights with the lowest magnitude (including any zero weights, e.g., non-active). Forcing this pattern on an existing unstructured mask might remove active (non-zero) weights, i.e., flipping some of the mask values from one to zero. We named those required flips, \textit{pattern-violations}. Removing active weights without re-training tends to severely degrade the model accuracy. To demonstrate the problem we used an unstructured sparse pretrained ResNet-50 model ($\rho=0.86$) and set the $N:M$ structure per-layer, based on \cref{eq:P_sparse}, such that the probability for a pattern-violation would be equal or less than a given percentage. Here we used a block size of $M=8$. Notably, without any optimization even a $1\%$ pattern-violation results in severe degradation (\cref{fig:violation_mean_adaprune}). Next, we describe two light methods to boost the accuracy.

\paragraph{Fixing pruning bias using mean absorption.} Several works \citep{banner2018post,finkelstein2019fighting,Hubara2020ImprovingPT} reported that it is important to fix the bias introduced when quantizing the model. We build on those results and suggest absorbing the mean of the $N$ pruned weights into the $M-N$ non zeroed weights. As can be seen in \cref{fig:violation_mean_adaprune} this simple fix, by itself, greatly boosts accuracy.
\vspace{-2mm}
\paragraph{AdaPrune.} Recently, several works \citep{Hubara2020ImprovingPT,markus2020} suggested light and fast fine-tuning techniques for post-train quantization. These techniques replace the heavy full model training with a fast per-layer optimization which requires only a few iterations to converge. While each method applies a different optimization technique, they all aim to reduce the discrepancy between the quantized and full-precision layer outputs. We adjusted the parallel-AdaQuant \citep{Hubara2020ImprovingPT} technique to the pruning problem by defining its objective to be:
\begin{equation}
    \min_{W'} ||WX- (S\odot W') X||_2^2 ,
\end{equation}
where $W$ is the original weight layer, $W'$ is the weight layer we aim to find, $X$ is the output of the previous activation layer, $S$ is the weight sparsity mask, and $\odot$  is a component-wise product. We named this method AdaPrune. In our experiments we used 1000 images from the ImageNet training set as a calibration set. As can be seen in \cref{fig:violation_mean_adaprune}, AdaPrune is capable of correcting the remaining error and obtain less than 1\% degradation from the original unstructured-sparse model counterpart. We argue that with AdaPrune, we can potentially adapt any generic mask to the hardware at hand, thus elevate the need to retrain the model. However, full re-training is still necessary when starting from a dense model (thus having 50\% pattern-violation), since there we get 2.3\% degradation using AdaPrune. We discuss and extend those experiments in \cref{secApp:exp}. 
\vspace{-2mm}
\section{Conclusions}
\label{sec:conclusions}
In this work, we analyze the constraints introduced by fine-grained sparsity. We discuss the limitations of current research in the field and suggest a new measure called \textit{mask diversity} to connect between the mask constraint and the accuracy of the network. In addition, we discuss the inherent problem of accelerating sparse training and suggest a novel N:M transposable-mask which enables accelerating the backward phase as well. We formulate the question of finding the optimal mask as a minimum-cost-flow problem and show no accuracy degradation in a variety of tasks with 2x acceleration in comparison to previous methods \citep{Nvidia}. We experimented with transposable-masks also in the setting of \citet{n:mStructured} where pretrained model is not given but the mask is dynamically changing based on the gradients and a copy of the weights which are kept dense. In this setting we define an approximation scheme with an (almost) linear (in input-size) time complexity that produces a mask whose $\ell_1$-norm is within a factor of 2 from the optimal mask and allow 2x acceleration in comparison to \citet{n:mStructured}. While a different line of research  suggested more extreme setting which restrict the memory footprint to the compressed model size \cite{bellec2017deep,mocanu2018scalable,mostafa2019parameter,dettmers2019sparse,evci2020rigging}, we argue that our method is orthogonal to it and both methods could be easily be combined. We believe this work paves the path toward true efficient sparse training. Finally, we suggest a simple method (AdaPrune) to transform an unstructured sparse model to an $N:M$ fine-grained sparse structured model with little to no training (e.g., less than 1\% degradation in ResNet50).  Furthermore, with a light training procedure over a calibration set (i.e., AdaPrune) we can switch from unstructured model to 4:8 structure sparsity and thus gain 2x inference speedup. 

\textbf{Broader impact.} Training DNNs is an expensive process which requires a high amount of resources and time. This long time can prevent the use of DNNs in many applications, despite their impressive performance. Reducing training time gives the opportunity to to use of DNNs in additional applications, or even use the extra time to improve the existing models. We need to take into account that accelerated training could introduce training instabilities. These models will need a careful examination, specially when used in real applications, such as medical devices.
 
\section*{Acknowledgements}
The research of JN is  supported in part by US-Israel BSF grant 2018352 and by ISF grant 2233/19 (2027511). The research of DS was supported by the Israel Science Foundation (grant No. 1308/18), and by the Israel Innovation Authority (the Avatar Consortium).

\bibliography{main}
\bibliographystyle{neurips}

\section*{Checklist}

The checklist follows the references.  Please
read the checklist guidelines carefully for information on how to answer these
questions.  For each question, change the default \answerTODO{} to \answerYes{},
\answerNo{}, or \answerNA{}.  You are strongly encouraged to include a {\bf
justification to your answer}, either by referencing the appropriate section of
your paper or providing a brief inline description.  For example:
\begin{itemize}
  \item Did you include the license to the code and datasets? \answerYes{See Section~\ref{gen_inst}.}
  \item Did you include the license to the code and datasets? \answerNo{The code and the data are proprietary.}
  \item Did you include the license to the code and datasets? \answerNA{}
\end{itemize}
Please do not modify the questions and only use the provided macros for your
answers.  Note that the Checklist section does not count towards the page
limit.  In your paper, please delete this instructions block and only keep the
Checklist section heading above along with the questions/answers below.

\begin{enumerate}

\item For all authors...
\begin{enumerate}
  \item Do the main claims made in the abstract and introduction accurately reflect the paper's contributions and scope?
    \answerYes{The main contribution of this paper include answering the three questions presented in the introduction and summarize in \cref{fig:chart}}
  \item Did you describe the limitations of your work?
    \answerYes{As explained in \cref{sec:conclusions}, there can be models that we didn't check where the proposed method could introduce training instabilities. We didn't notice any instability in all the models we checked. \BCH{?}}
  \item Did you discuss any potential negative societal impacts of your work?
    \answerYes{Yes, in section \cref{sec:conclusions} in broader impact we discuss the positive social impact of this work - reduce training time and allow the use of DNNs in additional applications. We don't see any negative impact of this work.}
  \item Have you read the ethics review guidelines and ensured that your paper conforms to them?
    \answerYes{}
\end{enumerate}

\item If you are including theoretical results...
\begin{enumerate}
  \item Did you state the full set of assumptions of all theoretical results?
    \answerYes{See \cref{app:lema proof}}
	\item Did you include complete proofs of all theoretical results?
    \answerYes{See \cref{app:lema proof}
}
    \end{enumerate}

\item If you ran experiments...
\begin{enumerate}
  \item Did you include the code, data, and instructions needed to reproduce the main experimental results (either in the supplemental material or as a URL)?
    \answerYes{As written in the abstract - A reference implementation can be found at \url{https://github.com/papers-submission/structured_transposable_masks}. }
  \item Did you specify all the training details (e.g., data splits, hyperparameters, how they were chosen)?
    \answerYes{See \cref{sec:appExpSettings}}
	\item Did you report error bars (e.g., with respect to the random seed after running experiments multiple times)?
    \answerNo{The experiments were run with random seed one time, as in the previous works \cite{n:mStructured,Nvidia} we compare to. Error bar not applicable.}
	\item Did you include the total amount of compute and the type of resources used (e.g., type of GPUs, internal cluster, or cloud provider)?
    \answerYes{See \cref{sec:appExpSettings}}
\end{enumerate}

\item If you are using existing assets (e.g., code, data, models) or curating/releasing new assets...
\begin{enumerate}
  \item If your work uses existing assets, did you cite the creators?
    \answerYes{In the url  \url{https://github.com/papers-submission/structured_transposable_masks} that includes our implementation, we cite the creator of existing assets with link to their license }
  \item Did you mention the license of the assets?
        \answerYes{In the url  \url{https://github.com/papers-submission/structured_transposable_masks} that includes our implementation, we cite the creator of existing assets with link to their license }
  \item Did you include any new assets either in the supplemental material or as a URL?
        \answerYes{As written in the abstract - A reference implementation can be found at \url{https://github.com/papers-submission/structured_transposable_masks}. }
  \item Did you discuss whether and how consent was obtained from people whose data you're using/curating?
    \answerNA{}
  \item Did you discuss whether the data you are using/curating contains personally identifiable information or offensive content?
   \answerNA{}
\end{enumerate}

\item If you used crowdsourcing or conducted research with human subjects...
\begin{enumerate}
  \item Did you include the full text of instructions given to participants and screenshots, if applicable?
    \answerNA{}
  \item Did you describe any potential participant risks, with links to Institutional Review Board (IRB) approvals, if applicable?
    \answerNA{}
  \item Did you include the estimated hourly wage paid to participants and the total amount spent on participant compensation?
    \answerNA{}
\end{enumerate}

\end{enumerate}


\newpage

\appendix

\renewcommand\thefigure{\thesection.\arabic{figure}} 
\renewcommand\thetable{\thesection.\arabic{table}} 
\renewcommand\theequation{\thesection.\arabic{equation}}  
\setcounter{figure}{0}  
\setcounter{table}{0}
\setcounter{equation}{0}

\crefalias{section}{appsec}
\crefalias{subsection}{appsec}
\crefalias{subsubsection}{appsec}

\section{Supplementary Material}
\subsection{Additional AdaPrune experiments}
\label{secApp:exp}

To further examine AdaPrune capabilities we checked two additional settings: (a) starting from pre-trained dense model, and (b) staring from less constrained N:M mask.
\subsubsection{AdaPrune from Dense}
While this case is more common, we expect to see some degradation as we know that we have 50\% mask violations. Yet as can be seen in table 2 we managed to restore accuracy to 2-3\% of the full-precision baseline using just AdaPrune. To further improve results we applied batch-norm-tuning as suggested by \citet{Hubara2020ImprovingPT} and kept the first and last layers dense which results in less than 2\% degradation. We believe it to be the first tolerable post-training-pruning results reported.

\begin{table}[h]
\centering
\caption{Using AdaPrune from dense pre-trained model. AP stands for AdaPrune and BNT stands for batch-norm-tuning.}
\label{tab:adaprune_dense}
\begin{tabular}{l|l|c|c|c}
\toprule
\multirow{2}{*}{Model}  & \multirow{2}{*}{Dense} & BiasFix & AP & AP\\ 
 & & BNT&  &  BNT\\ \hline
ResNet18 & 69.7\%& 62.47\%& 68.41\% & 68.63\% \\  \hline
ResNet34 &  73.3\% & 68.72\% & 72.15\% & 72.36\% \\  \hline
ResNet50 &  76.1\%& 67.42\% & 74.41 & 74.75\% \\ \hline
ResNet101 &  77.27 \%& 71.54\% & 76.36\% & 76.48\% \\  \midrule

\end{tabular}
\end{table}

\subsubsection{AdaPrune from N:M sparse}
In \cref{sec:mask_diversity} we explained why as the block size decreases the mask diversity decreases. Thus, we expect to have many violations when a pre-trained sparse model with $N_1:M_1$ translates to $N_2:M_2$, for $N_1 > N_2$ and $M_1 > M_2$. We argue that this might be a common case in the future as different hardware vendors would support different formats. In table \cref{tab:adaprune_sparse} we can see results of converting ResNet-50 model trained with $4:8$ sparsity pattern to $2:4$ and $1:2$ patterns. As can be seen, converting from $4:8$ to $2:4$ produces results with negligible accuracy degradation (less than 0.5\%). Therefore, we argue that  AdaPrune is an efficient and useful approach to convert models which were optimized on a different hardware than the one in use, as it removes the need for full sparse training. This is even more important when the training data is not available. 

\begin{table}[h]
\centering
\caption{Using AdaPrune to convert from one sparse pattern to the other. The baseline model was trained with 4:8 sparsity (90 epochs). Thus, 4:8 column is the baseline. BNT stands for batch-norm-tuning}
\label{tab:adaprune_sparse}
\begin{tabular}{l|c|c|c|c|c}
\toprule
\multirow{2}{*}{Model} & 4:8 & 2:4 & 2:4 &1:2 & 1:2 \\ 
 & & & BNT & & BNT \\ \hline
RN50 & 76.5\% & 76.2\% &76.4\%& 74.6\% & 75.1\% \\\hline
RN50-T & 77.1\% & 76.3\% &76.4\%& 74.7\% & 75.1\% \\\hline
RN18-T & 70.75\% & 70.1\% &70.2\%& 68.9\% & 69.2\% \\\midrule
\end{tabular}
\end{table}

\subsection{Proof of Lemma}
\label{app:lema proof}
\begin{theorem}
Algorithm 1 produces a tight 2-approximate solution, i.e., $W(P)<2\cdot W^{*}$.
\end{theorem}

\begin{proof}
Consider any node $i\in V \setminus \{s,t\}$. Let $E'(i)=\{e'_1, e'_2, e'_3,...e'_{M/2} \}$ denote the edges of an optimal solution that are adjacent to node $i$ and sorted in ascending order from light to heavy. Let $E(i)=\{e_1, e_2, e_3,...e_{M/2} \}$ denote the first $M/2$ edges  adjacent to $i$ in $P$ with respect to the order in which \cref{alg:approx} picked them.   By construction, we have that for all edges in $E(i)$: 
\begin{equation}
    w(e_1)\leq w(e_2) ...\leq w(e_{\frac{M}{2}}).
\end{equation}
We note that we can truncate the list of $i$ at $M/2$, since if $i$ has more than $M/2$ edges adjacent to it in $P$, then any such edge $(i,j)$ would also appear in $E(j)$ (among the first $M/2$ edges adjacent to $j$) by the minimality of the solution $P$.  Thus, the union of the lists $E(i)$ contains all edges in $P$.
We now prove by induction  that for any $n$, $n\geq 1$,
\begin{equation}
w(e_n)\leq w(e'_n).   
\end{equation}
\begin{itemize} 
\item Base case ($n=1$): $w(e_1) \leq w(e'_1)$, since by construction of \cref{alg:approx}, edge $e_1$ is the lightest edge adjacent to node $i$.
\item Inductive step: assume $w(e_n) \leq w(e'_n)$, then it must hold that $w(e_{n+1}) \leq w(e'_{n+1}))$; otherwise, if $w(e_{n+1}) > w(e'_{n+1}))$, then $e'_{n+1}$ should have been considered before $e_{n+1}$ and also chosen by \cref{alg:approx}.
\end{itemize}
Thus, $$\sum_{j=1}^{M/2} w(e_j) \leq \sum_{j=1}^{M/2} w(e'_j).$$
To complete the proof, our goal is to charge the weight of the edges in $P$ to the weight of the edges in the optimal solution based on the above inequality. However, note that an edge $(i,j) \in P$ may appear in only one of the lists $E(i)$ or $E(j)$.
Thus, for example, two edges in $P$, $(i,j)$ and $(i',j)$, may charge their weight to the same edge $(i,i')$ in the optimal solution. Clearly, this ``double" charging can happen at most twice for each edge in the optimal solution, hence:
\begin{equation*}
W(P) \leq 2 W^{*}.
\end{equation*}
\end{proof}

\begin{figure*}[]
\begin{center}
\begin{subfigure}[]{0.3\linewidth}
\includegraphics[width=\columnwidth]{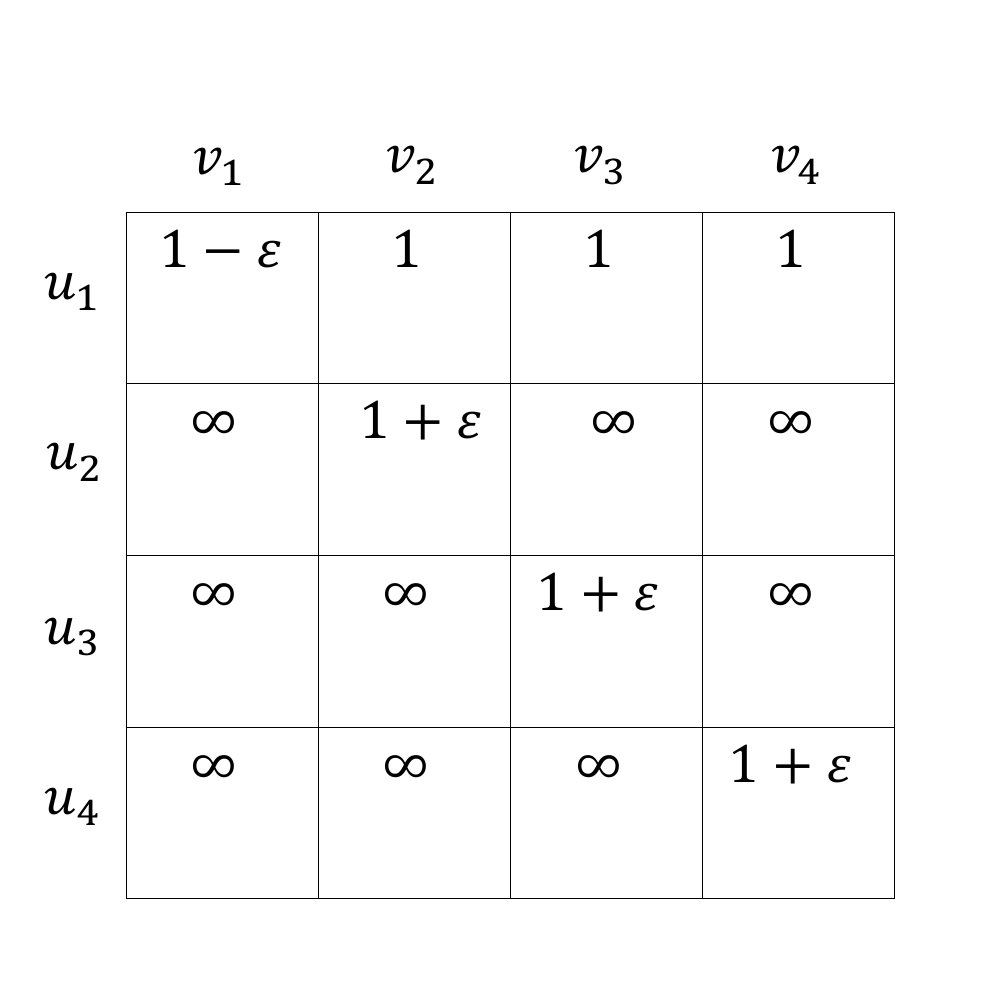}
\caption{}
\label{fig:tightmatrix}
\end{subfigure}
\hfill
\begin{subfigure}[]{0.3\linewidth}
\includegraphics[width=\columnwidth]{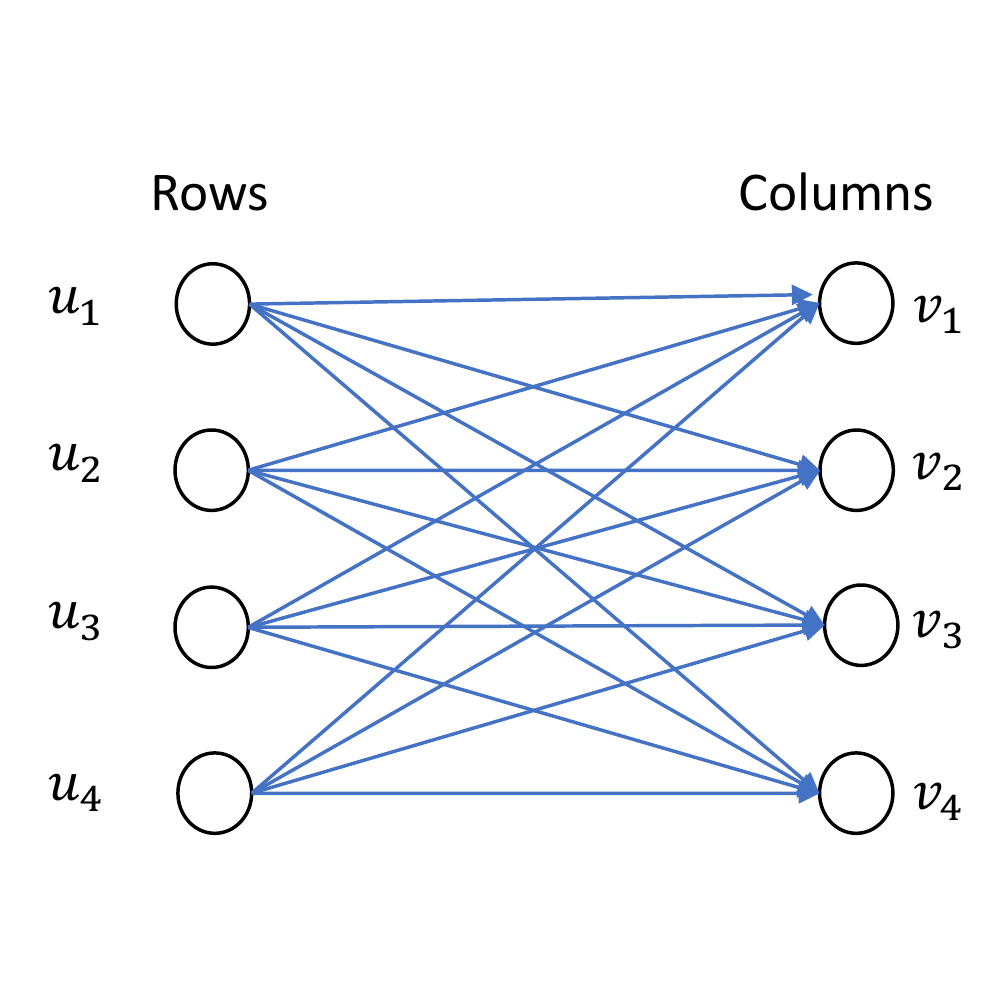}
\caption{}
\label{fig:tightgraph}
\end{subfigure}
\begin{subfigure}[]{0.3\linewidth}
\includegraphics[width=\columnwidth]{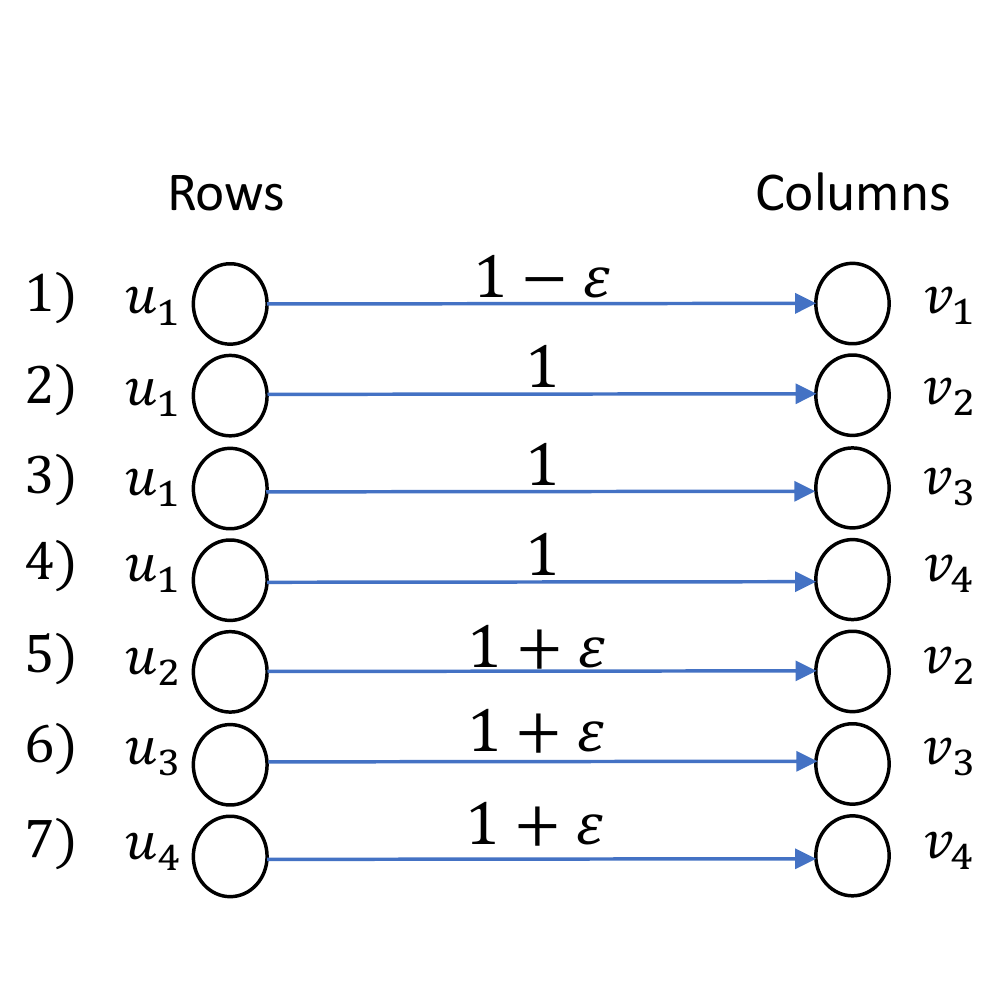}
\caption{}
\label{fig:tightiter}
\end{subfigure}
\caption{\textbf{(a):} Block of size $4 \times 4$ where we want to zero out one element in each row and column using the 2-approximation algorithm.   \textbf{(b):} The block represented as a directed bipartite graph.  \textbf{(c):} The 7 iterations of the 2-approximation algorithm on the bipartite graph. Notice that we get an approximation ratio of $\frac{7}{4}$, since the optimal solution picks only the diagonal entries.  
}
\end{center}
\end{figure*}

In the following, we show that our analysis of the upper bound of 2 on the approximation factor (proved in the lemma) is asymptotically tight. Consider the example in \cref{fig:tightiter}. Let us assume we want to zero out one element in each row and column in the block of size $4 \times 4$ presented in \cref{fig:tightmatrix} using the 2-approximate algorithm (\cref{alg:approx}). First, we need to convert the block into a bipartite graph (as suggested in Figure \ref{fig:minCost}). This construction appears in \cref{fig:tightgraph}. Next, we sort the the edges from light to heavy and go over the sorted list. In \cref{fig:tightiter} we show the seven iterations of the 2-approximate algorithm. All edges are added to the list of chosen edges $P$ up until the 7th iteration. The algorithm stops at the 7th iteration, since after adding edge $u_4 \xrightarrow{1} {v_4}$, every node is already ``covered" by at least one edge (in other words, each row and each column has at least one entry chosen for pruning). Note that the optimal solution would choose the edges that correspond to the entries on the diagonal (i.e.,  $u_1 \xrightarrow{1} {v_1},u_2 \xrightarrow{1} {v_2},u_3 \xrightarrow{1} {v_3}$, and $u_4 \xrightarrow{1} {v_4}$), summing up to a total weight of 4. Hence, we get an approximation ratio of $\frac{7}{4}$. It is easy to see that when using the same construction for a general block of size $M \times M$, we get an approximation ratio of $\frac{2M -1}{M}$, asymptotically converging to 2 as $M$ goes to infinity.

\subsection {Min cost flow vs. 2-approximation run-time analysis} 
\label{sec:appRunTime}
In this section we specify the complexity of different min-cost flow methods and compare them with the running time of our 2-approximation method. \citet{ahuja1988network} specifies the running times of six min-cost flow algorithms, two of which have distinctively better performance for our construction compared to the others (see \citet{ahuja1988network}[pp. 396-397]). The running time  of these two methods depends on the following parameters: number of nodes $n$, number of edges $m$, the largest weight coefficient $W$, and the flow demand $U$. 
The Goldberg-Tarjan algorithm and the double scaling algorithm have running times of
$\tilde{O}(mn)$, where $\tilde{O}$ hides polylogarithmic factors.
Thus, in our construction, for a block of size $M \times M$ the number of edges is $M^2+2M$, the number of nodes is $2M+2$, and the flow demand is $0.5M^2$. This boils down to running times of $\tilde{O}(M^3)$
for finding an optimal mask.
The running time of the 2-approximation algorithm in comparison is $\tilde{O}(M^2)$. \revision{ In \cref{running_resnet5024} we show the running time overhead of ResNet50 training using 2:4 transposable mask with IP, min cost flow and 2-approximation algorithms over regular training.  These overheads are smaller than what we measured for 4:8 (\cref{running_resnet50}). These methods were implemented in a non-optimized way, therefore, we expect a further decrease of the overhead in the 2-approximation in an optimized algorithm. }

\begin{table}[h]

\centering
\caption{\small{Overhead of different algorithms for finding the 2:4 transposable mask: ratio of their running time over regular training (ResNet50). Notice the overhead reduction in the 2-approximation algorithm in comparison to the naive IP.}}
\label{running_resnet5024}
\begin{tabular}{l|c}
\toprule
 Method &  Overhead (\%)   \\ \hline
Integer-programming  & 160 \\ \hline
Min-cost flow  &  65 \\ \hline
2-approximation & 13 \\ \bottomrule
\end{tabular}
\end{table}

\subsection{Experiments Setting}
In all our experiments we use 8 GPU GeForce RTX 2080 Ti.
\label{sec:appExpSettings}
\paragraph{AdaPrune} We used a small calibration set of 1000 images (one per-class). We run AdaPrune for 1000 iterations with batch-size of 100. For the results in the supplementary material, we kept the first and last layers dense.
\paragraph{N:M transposable sparsity mask from a pre-trained model}
We used torchvison \cite{torchvision} model-zoo as our pre-trained dense baseline. For all ResNet models we used the original regime as given by \citet{resnet}, i.e., SGD over 90 epochs starting with learning rate of 0.1 and decreasing it at epochs 30,60,80 by a factor of 10. For BERT-large and MaskRCNN we used the defaults scripts as in \citet{NvidiaTensorCores}.
\paragraph{N:M transposable sparsity mask from scratch}
We use the exact same setting as given by \citet{n:mStructured}.

\subsection{N:M hardware requirements}
\label{sec:appHW}
For conventional hardware, Broadly, N:M sparsity requires adding an N+1 to 1 multiplexer (N+1:1)to the adder tree in the code of the matrix multiplication. engine. Thus switching from 2:4 fine grained sparsity to 4:8 requires 5:1 multiplexers instead of 3:1. The simplest implementation of a multiplexer is build of set of 2:1 multiplexers which means that the area required for the multiplexers scale logarithmicly with the number of zeros in the block (N).

\subsection{Mask diversity Derivation}
\label{app:MD_deriv}
Let us consider $W$ to be a block of size $n \times n$ from a weight tensor and our desired sparsity level to be $N/M$. 
Thus, the MD of unstructured sparsity consists of all possibilities to pick $N$ values out of $B$ (\cref{eq:mask_d} (a)). The MD increases with the block size ($B$), which might explain the recent success of global pruning. Next we investigate, fine-grained $N:M$ structured sparsity \citep{Nvidia}. This approach requires us to zero out $N$ values in each block of size $M$. Since we have $\frac{T}{M}$ blocks this results in \cref{eq:mask_d} (b). If we wish to enforce the constraints on both row and columns of the matrix (i.e., $N:M$ transposable structured) the diversity decreases. Let us first assume $N=1$. The number of possibilities in each block of size $M^2$ is $M!$. Repeating this process for general $N$ in all the $\frac{B}{M^2}$ blocks results in \cref{eq:mask_d} (c). A more constrained mask, is a fine-grained $N:M$ mask with a sequential structure. Here we require that each $M$ contiguous elements would contain $N$ sequential zeros. In each block of size $M^2$, there are $M-N+1$ options of sequential zeros. Applying it on all the $\frac{B}{M}$ blocks results in \cref{eq:mask_d}(d).
\subsection{Mask diversity Experiments}
\label{app:MD_exp}
\begin{figure*}[h!]
\vspace{-2mm}
\begin{center}
\includegraphics[width=0.6\columnwidth]{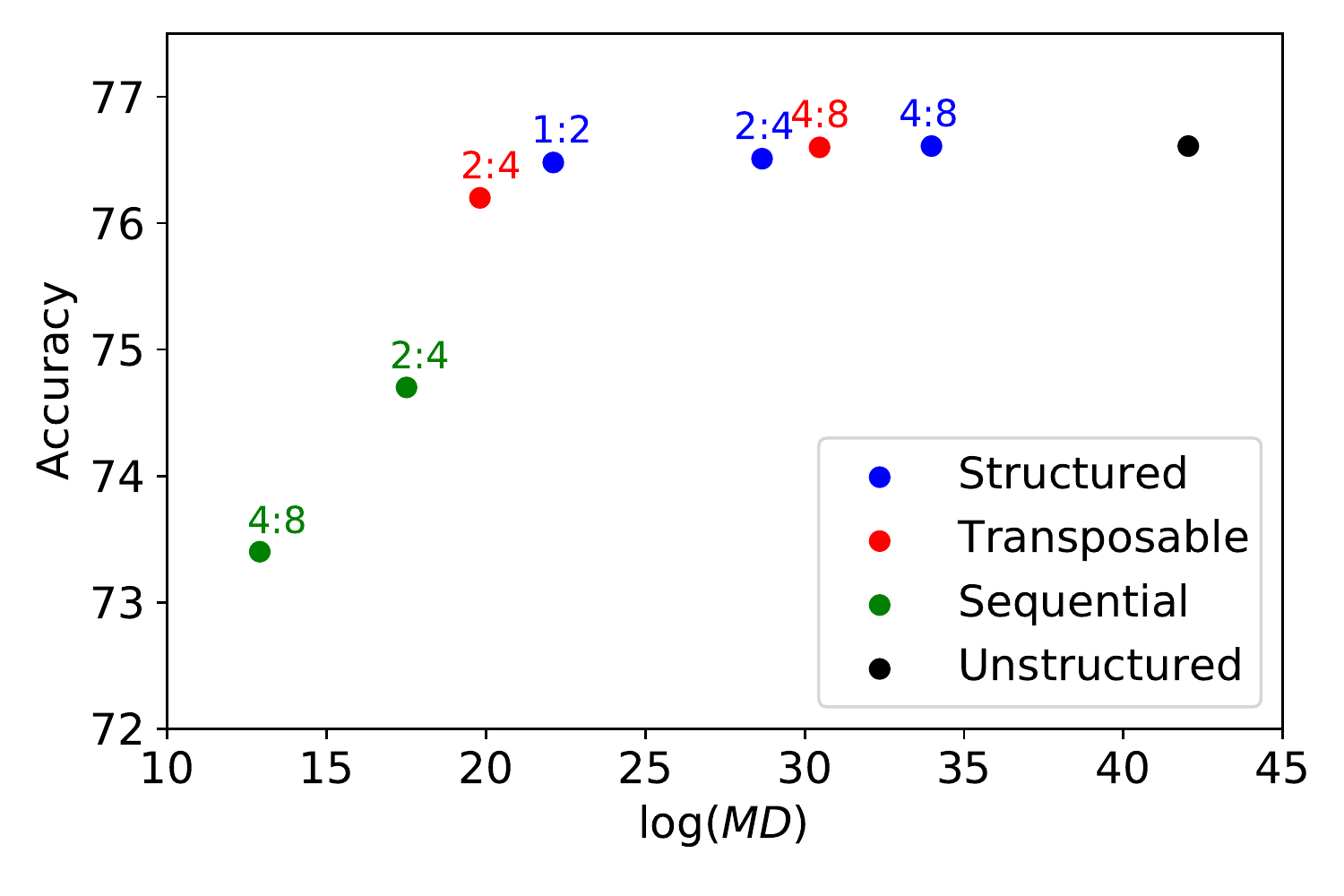}
\caption{}
 \label{fig:diversityResNet50}
\caption{\small{ResNet50 over ImageNet top-1 accuracy for weight sparsity of 50 \% using different structured and unstructured masks. As expected the mask diversity correlates with the pruned model accuracy.}}
\end{center}
\end{figure*}

In additional to the results in \cref{sec:mask_diversity} we experimented in \cref{fig:diversityResNet50} with ResNet50 over ImageNet dataset. In all our experiments we used one-shot pruning from dense model and applied the same regime as the original training as suggested by \citet{Nvidia}.

\subsection{2:4 transposable mask}
\label{sec:AppExp24}
\revision{In \cref{tab:App24} we show experiments with the 2:4 transposable mask in the "training from scratch" setting. Moreover, after we published the first version of our paper, researchers from NVIDIA \citep{Stosic2021SearchSF} continued our work and demonstrated on a large set of models that one can achieve less than 1\% degradation even with 2:4 transposable masks in the "training from a trained dense model" setting. As can be seen, 2:4 transpose mask can achieve high accuracy in part of the models. This results correlates with the shown MD, since 2:4 transpose mask has similar MD to 1:2 mask which already achieved high accuracy (\cref{fig:diversityResNet50}). Despite this, in some scenarios 2:4 transposable does not work as well, as in the case of finetuning BERT-Large on SQuAD dataset. Here the 2:4 transposable mask incurred a $\sim 1\%$ degradation (90.18 F1 vs. 91.1 F1 for the dense model) while a 4:8 transposable mask incurred less than 0.5\% degradation in F1 score (90.65 F1).  }

\begin{table}[h]
\centering
\caption{\small{Training from scratch of ResNet18, ResNet50 on imageNet dataset and fine-tuning of Bert-Large on SQuAD dataset with 2:4 transposable mask using the proposed 2-approximation scheme. SU refers to the sparse tensor cores utilization, used both in forward and backward passes in the proposed method, allowing 2x speedup in comparison to previous methods \citep{n:mStructured}}}
\label{tab:App24}
\begin{tabular}{l|l|c|c|c|c}
\toprule
\multirow{2}{*}{Model} & \multirow{2}{*}{Metric} & \multicolumn{2}{c|}{Baseline}  & \multicolumn{2}{c}{Ours 2:4-T} \\ \cline{3-6} 
 &  & \multicolumn{1}{c|}{SU} & \multicolumn{1}{c|}{Accuracy}  & \multicolumn{1}{c|}{SU} & \multicolumn{1}{c}{Accuracy} \\ \hline
ResNet18 & Top1 & \multirow{3}{*}{0\%} & 70.54\%  & \multirow{3}{*}{66\%} & 70.5\% \\ \cline{1-2} \cline{4-4} \cline{6-6} 
ResNet50 & Top1 &  & 77.3\%  &  & 77.1\% \\ \cline{1-2} \cline{4-4} \cline{6-6} 
Bert-Large & F1 &  & 91.1\% & & 90.18\% \\  \bottomrule
\end{tabular}
\end{table}

\end{document}